\documentclass[conference]{IEEEtran}

\usepackage{cite}
\usepackage{enumitem}
\usepackage{amsmath,amssymb,amsfonts}
\usepackage{graphicx}
\usepackage{textcomp}
\usepackage{xcolor}
\usepackage{placeins}
\usepackage{algorithm}
\usepackage[noend]{algpseudocode}
\usepackage{url}

\def\BibTeX{{\rm B\kern-.05em{\sc i\kern-.025em b}\kern-.08em
    T\kern-.1667em\lower.7ex\hbox{E}\kern-.125emX}}

\title{STROOBnet Optimization via GPU-Accelerated Proximal Recurrence Strategies}
\author{\IEEEauthorblockN{(Ted) Edward Holmberg}
\IEEEauthorblockA{\textit{Department of Computer Science} \\
\textit{University of New Orleans}\\
New Orleans, LA, USA \\
eholmber@uno.edu}
\and
\IEEEauthorblockN{Mahdi Abdelguerfi}
\IEEEauthorblockA{\textit{Cannizaro-Livingston Gulf States} \\
\textit{Center for Environmental Informatics}\\
New Orleans, LA, USA\\
mahdi@cs.uno.edu}
\and
\IEEEauthorblockN{Elias Ioup}
\IEEEauthorblockA{\textit{Center for Geospatial Sciences} \\
\textit{Naval Research Laboratory,}\\
\textit{Stennis Space Center}\\
Mississippi, USA. \\
elias.ioup@nrlssc.navy.mil}
}

\IEEEoverridecommandlockouts
\IEEEpubid{\footnotesize\begin{minipage}{\textwidth}\ \\[12pt]
\hrule width 0.49\columnwidth height 0.5pt \vspace{10pt}
Accepted version. \copyright 2023 IEEE. Personal use of this material is permitted. \\
Permission from IEEE must be obtained for all other uses.\\
DOI 10.1109/BigData59044.2023.10386774
\end{minipage}}

\begin{document}

\maketitle

\begin{abstract}
Spatiotemporal networks' observational capabilities are crucial for accurate data gathering and informed decisions across multiple sectors. This study focuses on the Spatiotemporal Ranged Observer-Observable Bipartite Network (STROOBnet), linking observational nodes (e.g., surveillance cameras) to events within defined geographical regions, enabling efficient monitoring. Using data from Real-Time Crime Camera (RTCC) systems and Calls for Service (CFS) in New Orleans, where RTCC combats rising crime amidst reduced police presence, we address the network's initial observational imbalances. Aiming for uniform observational efficacy, we propose the Proximal Recurrence approach. It outperformed traditional clustering methods like k-means and DBSCAN by offering holistic event frequency and spatial consideration, enhancing observational coverage.
\end{abstract}

\begin{IEEEkeywords}
Bipartite Networks, Spatiotemporal Networks, Observer-observable Relationships, Proximal Recurrence, Network Optimization
\end{IEEEkeywords}

\section{Introduction}
In response to escalating crime and a diminished police presence \cite{SimermanAdelson2022}, New Orleans—besieged by over 250,000 Calls for Service (CFS) in 2022, of which over 13,000 were violent—has adopted Real-Time Crime Camera (RTCC) systems as a strategic countermeasure\cite{CityOfNewOrleans2022}. The RTCC, with 965 cameras saved over 2,000 hours of investigative work in its inaugural year\cite{Genetec2019}. However, a significant challenge persists: optimizing these systems to maximize their efficacy and coverage amidst the city's staggering crime rate.

This research focuses on the mathematical model of this issue through the lens of graph theory, exploring Spatiotemporal Ranged Observer-Observable Bipartite Networks (STROOBnets). While the immediate application is crime surveillance, the principles and methodologies developed herein hold applicability to various observer-type networks, including environmental monitoring systems.

The objectives are threefold: identify the most influential nodes, evaluate network efficacy, and enhance network performance through targeted node insertions. Employing a centrality measure, the methodology optimizes observer node placements in a spatial bipartite network. Through modeling relationships between service calls, violent events, and crime camera locations in New Orleans using knowledge graphs, the research contrasts our Spatiotemporal Ranged Observer-Observable clustering technique with alternative approaches, such as Kmeans\cite{{wu2021kmeans}}, DBSCAN \cite{schubert2017dbscan}, and time series analysis via statistical mode \cite{hastie2001}. Additionally, our approach, focusing on GPU-accelerated computation, ensures scalable and efficient data processing across large datasets \cite{rapids2023}.

\section{Background}
This section outlines the key datasets and mathematical concepts foundational to our exploration of spatiotemporal clustering in crime and surveillance data.

\subsection{Dataset: Crime Dynamics and Surveillance in New Orleans}
This subsection introduces two datasets used to construct the STROOBnet.

\subsubsection{Real-Time Crime Camera (RTCC) Systems}
The RTCC system, set up by the New Orleans Office of Homeland Security in 2017, uses 965 cameras to monitor, respond to, and investigate criminal activity in the city. Of these cameras, 555 are owned by the city and 420 are privately owned \cite{Stein2021}.

\subsubsection{Calls for Service (CFS)}
The New Orleans Police Department’s (NOPD) Computer-Aided Dispatch (CAD) system logs Calls for Service (CFS). These include all requests for NOPD services and cover both calls initiated by citizens and officers \cite{CityOfNewOrleans2022}.

\subsection{Mathematical Background}
Utilizing the datasets introduced, our analysis hinges upon several mathematical and network principles, each playing a crucial role in interpreting the spatial and temporal dynamics of crime and surveillance in New Orleans, Louisiana (NOLA).

\vspace{6pt}
\subsubsection{Bipartite Networks}
Bipartite networks consist of two discrete sets of nodes with edges connecting nodes from separate sets. They effectively portray observer-observable relationships by defining the relationship between the two types.

Mathematically, a bipartite graph (or network) \( G = (U, V, E) \) is defined by two disjoint sets of vertices \( U \) and \( V \), and a set of edges \( E \) such that each edge connects a vertex in \( U \) with a vertex in \( V \). Formally, if \( e = (u, v) \) is an edge in \( E \), then \( u \in U \) and \( v \in V \). This ensures that nodes within the same set are not adjacent and that edges only connect vertices from different sets \cite{asratian1998}.

\vspace{6pt}
\subsubsection{Centrality}
Degree Centrality measures a node's influence based on its edge count and is crucial for identifying critical nodes within a network \cite{Vincenzo2013}. In bipartite networks, such as those modeling interactions between surveillance cameras (observers) and crime incidents (observables), centrality is vital for evaluating and optimizing camera positioning to ensure effective incident monitoring. Consequently, the degree centrality of the observer nodes is of particular concern.

\vspace{6pt}
\subsubsection{Spatiotemporal Networks}
Spatiotemporal networks (STNs) model entities and interactions that are both spatially and temporally situated. These networks can effectively represent geolocated events and observers by establishing connections based on spatiotemporal conditions. 

A Spatiotemporal Network (STN) is defined as a sequence of graphs representing spatial relationships among objects at discrete time points:
\begin{equation}
STN = \Big( G_{t_1}, \; G_{t_2}, \; ..., \;  G_{t_n} \Big) 
\end{equation}
where each graph \( G \) is defined as:
\begin{equation}
G_{t_i} = \Big( N_{t_i}, \; E_{t_i} \Big)
\end{equation}
and represents the network at time \( t_i \) with \( N_{t_i} \) and \( E_{t_i} \) denoting the set of nodes and edges at that time, respectively. \cite{holme2012} 

In the context of this research, STNs are utilized to model relationships between crime cameras (observer nodes) and violent events (observable nodes) across New Orleans. Nodes represent objects, whereas edges depict spatiotemporal relationships, connecting observer nodes to observable nodes based on spatial proximity and temporal occurrence. The analysis of STNs allows for the extraction of insightful patterns, aiding in understanding and potentially mitigating the progression and spread of events throughout space and time. 

\vspace{6pt}
\subsubsection{Spatiotemporal Clustering}
Spatiotemporal clustering groups spatially and temporally proximate nodes to identify regions and periods of significant activity within a network. In the context of spatiotemporal networks, clusters might reveal hotspots of activity or periods of unusual event concentration. \cite{Holmberg2022}  Techniques for spatiotemporal clustering must consider both spatial and temporal proximity, ensuring that nodes are similar in both their location and their time of occurrence\cite{ladner2002}.

\section{Related Work}
This section assesses prevalent clustering techniques, underscoring their drawbacks when applied to bipartite networks, especially in our context. The subsequent results section will contrast these methods with our proposed approach, emphasizing their performance in bipartite spatiotemporal networks scenarios.

\subsection{K-means Clustering}
K-means clustering segregates datasets into \( k \) clusters, minimizing intra-cluster discrepancies \cite{wu2021kmeans}. However, K-means poses challenges when applied to spatiotemporal data. It requires pre-specification of \( k \), which might not always be intuitive. The method's insistence on categorizing every data point can obfuscate pivotal spatial-temporal patterns. This can misrepresent genuine structures in observer-observable networks. Furthermore, K-means' inability to set a maximum cluster radius or diameter means it doesn't consider entities' observational range in the network, such as surveillance cameras. This hinders its usage where spatial influence is paramount, necessitating alternative clustering strategies that accommodate spatial limitations \cite{dorabiala2022spatiotemporal}.

\subsection{DBSCAN}
DBSCAN clusters based on density proximity and doesn't require specifying cluster numbers upfront \cite{schubert2017dbscan, birant2007stdbscan}. However, in bipartite spatiotemporal networks, DBSCAN can conflate adjacent clusters, forming larger, potentially sparser clusters. This can lead to misinterpretations of genuine data patterns. Additionally, DBSCAN's lack of generated centroids for clusters limits its potential for strategizing node placements, particularly where centroids represent efficient insertion points.

\subsection{Mode Clustering}
The statistical mode pinpoints recurrent instances in datasets, offering insights on frequent occurrences \cite{hastie2001}. However, its inherent limitation lies in overlooking spatial relationships. By focusing solely on frequency, mode clustering can neglect spatial clusters that, while not being the most frequent, play a critical role in the network. This oversight can lead to ineffective strategies for node placements, undermining the utilization of spatial-temporal patterns.

\section{Approach}

\subsection{Problem Definition}

Let matrices representing observer nodes \(O\) and observable events \(E\) be defined, with their respective longitudinal \((O_{\text{lon}}, E_{\text{lon}})\) and latitudinal \((O_{\text{lat}}, E_{\text{lat}})\) coordinates. The objective is to formulate a framework that:
\begin{itemize}
    \item Computes distances between observers and events.
    \item Assesses the centrality of observer nodes.
    \item Classifies events according to their observability.
    \item Clusters unobserved points utilizing spatial proximity.
    \item Adds new observers to improve network performance.
\end{itemize}

\subsection{Objectives}

\begin{enumerate}
\item \textbf{Maximize Current Network Observability:} Optimize the placement or utilization of existing observer nodes to ensure a maximum number of events are observed.
\item \textbf{Identify and Target Key Unobserved Clusters:} Analyze unobserved events to identify significant clusters and understand their characteristics to inform future observer node placements.
\item \textbf{Strategize Future Observer Node Placement:} Develop strategies for placing new observer nodes to address unobserved clusters and prevent similar clusters from forming in the future.
\end{enumerate}

\subsection{Rationale}

The limitations identified within existing methodologies, as discussed in the Related Work section, underscore the need for an innovative approach to optimizing bipartite networks, particularly in the context of spatiotemporal data. Traditional clustering methodologies, such as K-means and DBSCAN, present challenges in terms of accommodating spatial constraints, managing computational complexity, and providing actionable insights for node insertions. Whereas non-spatial methods like statistical mode lack the capacity to truly harness the spatial-temporal patterns within the data, often leading to suboptimal strategies for node insertions.

Therefore, our approach hinges on creating a bipartite distance matrix to systematically evaluate the spatial relationships between observer nodes and events. Following this, clustering algorithms are implemented to group disconnected points, providing an understanding of the spatial dimensions of our data. This methodology evaluates the existing observer network and delivers strategic, data-driven insights to enhance future network configurations.

\begin{itemize}
    \item \textbf{Distance Matrix Calculation:} Utilizing geographical data to calculate distances between observer nodes and events, with particular attention to ensuring all possible combinations of nodes and events are considered, is paramount to understanding spatial relationships within the network.
    \item \textbf{Effectiveness Evaluation:} The centrality and effectiveness of observer nodes are crucial metrics that inform us about the current status of the network in terms of its observational capabilities.
    \item \textbf{Event Classification:} Classifying events into observed and unobserved categories helps in understanding the coverage of the observer network and identifying potential areas of improvement.
    \item \textbf{Clustering of Unobserved Points:} Identifying clusters amongst unobserved points guides the strategic placement of new observer nodes, ensuring they are positioned where they can maximize their observational impact.
\end{itemize}

\subsection{Challenges}

\begin{itemize}
    \item \textbf{Computational Complexity:} Given the potentially large number of observer nodes and events, computational complexity is a pertinent challenge, especially when calculating the bipartite distance matrix and implementing clustering algorithms.
    \item \textbf{Spatial Constraints:} Ensuring that the placement of new observer nodes adheres to geographical and logistical constraints while still maximizing their observational impact. The range of each observer node is a major consideration for any event clustering.
    \item \textbf{Temporal Dynamics:} Accounting for the temporal dynamics within the data, ensuring that the models and algorithms are robust enough to handle variations and fluctuations in the event occurrences over time.
\end{itemize}

\section{Methods}
This section provides a detailed account of the approaches and algorithms used to establish spatiotemporal relationships between observer nodes and observable events. The ensuing subsections systematically unfold the mathematical and algorithmic strategies employed in various processes: calculating distances using the Haversine formula, determining centrality and generating links, constructing bipartite and unipartite distance matrices, classifying events, initializing STROOBnet, and clustering disconnected events. Each subsection introduces relevant notations, explains the method through mathematical expressions, and outlines the procedural steps of the respective algorithm.

~\\
\hrule width \columnwidth
\section*{Bipartite Distance Matrix}

Constructing a bipartite distance matrix is pivotal in capturing the spatial relationships between observer nodes and events. 

\textbf{Notations:}
\begin{itemize}
    \item \( O \): Matrix representing observer nodes with coordinates as its elements, where \( O_{ij} \) denotes the \( j \)-th coordinate of the \( i \)-th observer. Example: 
    \[
    O = \begin{bmatrix} x_1 & y_1 \\ x_2 & y_2 \\ \vdots & \vdots \end{bmatrix}
    \]
    \item \( E \): Matrix representing observable events with coordinates as its elements, where \( E_{ij} \) denotes the \( j \)-th coordinate of the \( i \)-th event. Example: 
    \[
    E = \begin{bmatrix} x'_1 & y'_1 \\ x'_2 & y'_2 \\ \vdots & \vdots \end{bmatrix}
    \]
    \item \( DM \): Distance Matrix, where \( DM_{ij} \) represents the distance between the \(i\)-th observer and the \(j\)-th event. Assuming there are \(m\) observers and \(n\) events, \( DM \) will be an \(m \times n\) matrix. Example: 
    \[
    DM = \begin{bmatrix} 
    d_{11} & d_{12} & \dots  & d_{1n} \\
    d_{21} & d_{22} & \dots  & d_{2n} \\
    \vdots & \vdots & \ddots & \vdots \\
    d_{m1} & d_{m2} & \dots  & d_{mn} \\
    \end{bmatrix}
    \]
    where each element \(d_{ij}\) represents the distance between the coordinates of the \(i\)-th observer and the \(j\)-th event.

\end{itemize}

\textbf{Methodology:}
\begin{enumerate}
    \item \textit{Haversine Distance Calculation:}
    Compute the Haversine distances for all combinations of observer-event pairs, populating the distance matrix \(DM\). Specifically, 
    \[
    DM_{ij} = \text{haversine}(O_i, E_j)
    \]
    where \( O_i \) and \( E_j \) are the coordinates of the \( i \)-th observer and \( j \)-th event. The Haversine formula calculates the shortest distance between two points on the surface of a sphere, given their latitude and longitude. The resulting distances in \(DM\) are given in kilometers, assuming the Earth's mean radius to be 6371 km.\cite{wilson2003, Chung2001}

    \vspace{6pt}
    \item \textit{Centrality Calculation:}
    The centrality of each observer is calculated as the count of events within a specified radius \( r \). This is mathematically expressed as:
    \[
    \text{Centrality}(o) = \sum_{e \in E} \mathbb{I}(DM_{oe} \leq r)
    \]
    In this equation, \( \mathbb{I} \) represents the indicator function, which is defined as:
    \[
    \mathbb{I}(P) = 
    \begin{cases} 
    1 & \text{if } P \text{ is true} \\
    0 & \text{if } P \text{ is false}
    \end{cases}
    \]
    Hence, \( \mathbb{I}(DM_{oe} \leq r) \) is 1 if the distance between observer \( o \) and event \( e \), denoted as \( DM_{oe} \), is less than or equal to \( r \), and 0 otherwise. The centrality thus provides a count of events that are within the radius \( r \) of each observer.

    \vspace{6pt}
    \item \textit{Link Generation:}
    Identify and create links between observers and observables that are within radius \( r \) of each other, forming a set \( L \) of pairs \((o, e)\) such that 
    \[
    L = \{(o, e) \,|\, DM_{oe} \leq r\}
    \]
    where each pair represents a link connecting observer \( o \) and event \( e \) in the bipartite graph, constrained by the specified radius.

\end{enumerate}

\textbf{Algorithm Overview:}
\begin{enumerate}
    \item Compute Haversine Distances for all observer-event pairs.
    \item Determine centrality of each observer based on events within radius \( r \).
    \item Generate links between observers and events within radius \( r \).
\end{enumerate}

\vspace{6pt}
\hrule width \columnwidth
\section*{Event Classifier}

\textbf{Notations:}
\begin{itemize}
    \item \( E \): Matrix representing events, as defined in the "Bipartite Distance Matrix" section.
    \item \( DM \): Distance Matrix, as defined in the "Bipartite Distance Matrix" section.
    \item \( r \): Radius threshold.
    \item \( OE \) and \( UE \): Sets of indices of observed and unobserved events.
\end{itemize}

\textbf{Methodology:}
\begin{enumerate}
    \item \textit{Number of Observations Calculation:}
    Calculate the number of observations for each event.
    \[
    num_{observations}(e) = \sum_{i=1}^{|O|} \mathbb{I}(DM_{ie} \leq r) \, \forall e \in E
    \]
    where \( \mathbb{I} \) is the indicator function that is 1 if the condition inside is true and 0 otherwise, and \( |O| \) is the number of observers.
    
    \item \textit{Determine Observed and Unobserved Points:}
    Classify the events into observed and unobserved based on the number of observations.
    \[
    \begin{aligned}
        OE &= \{ e \,|\, num_{observations}(e) > 0, \, e \in E \} \\
        UE &= \{ e \,|\, num_{observations}(e) = 0, \, e \in E \}
    \end{aligned}
    \]
\end{enumerate}

\textbf{Algorithm Overview:}
\begin{enumerate}
    \item Calculate the number of observations for each event.
    \item Classify events into observed (set \( OE \)) and unobserved (set \( UE \)).
\end{enumerate}

\vspace{6pt}
\hrule width \columnwidth
\section*{Initialize STROOBnet}

\textbf{Notations:}
\begin{itemize}
    \item \( O \) and \( E \): Sets of RTCC (observers) and CFS (events).
    \item \( L \): Set of Links between observers and events.
    \item \( r \): Radius threshold, consistent with prior sections.
    \item \( OE \) and \( UE \): Sets of indices of observed and unobserved events, maintaining consistency with the "Event Classifier" section.
\end{itemize}

\textbf{Methodology:}
\begin{enumerate}
    \item \textit{Bipartite Distance Calculation:}
    Compute the distances among RTCC nodes and CFS nodes.
    \[
    DM, L  = \text{Bipartite\_Distance\_Matrix}(O, E, r)
    \]
    
    \item \textit{Determine Observed and Unobserved Points:}
    Identify which events are observed and which are not.
    \[
    OE, UE = \text{Event\_Classifier}(E, DM, r)
    \]
    
\end{enumerate}

\textbf{Algorithm Overview:}
\begin{enumerate}
    \item Calculate distances among RTCC nodes and CFS nodes.
    \item Determine which events are observed and unobserved.
\end{enumerate}

\vspace{6pt}
\hrule width \columnwidth
\section*{Unipartite Distance Matrix}

\textbf{Notations:}
\begin{itemize}
    \item \( UE = \{e_1, e_2, \ldots, e_m\} \) be a set of unobserved event nodes.
    \item \( DM \) is a distance matrix, where \( DM_{ij} \) represents the distance between nodes \( e_i \) and \( e_j \).
    \item \( r \) is a radius threshold.
\end{itemize}

\textbf{Methodology:}
\begin{enumerate}
    \item \textit{Point Extraction:}
    Extract the longitudinal (\( x \)) and latitudinal (\( y \)) coordinates of unobserved nodes.
    \[
    P_x = \{x_1, x_2, \ldots, x_m\}, \quad P_y = \{y_1, y_2, \ldots, y_m\}
    \]

    \item \textit{Combination of Nodes:}
    Compute all possible combinations of nodes from \( UE \).
    \[
    C = \{(e_i, e_j) : e_i, e_j \in UE, i \neq j\}
    \]

    \item \textit{Haversine Distance Calculation:}
    Compute the haversine distance for each combination of nodes in \( C \) and construct the distance matrix \( DM \).
    \[
    DM_{ij} = \text{haversine}(e_i, e_j) \quad \forall (e_i, e_j) \in C
    \]
    where \(\text{haversine}(e_i, e_j)\) calculates the haversine distance between nodes \( e_i \) and \( e_j \).
    
    \item \textit{Distance Matrix Construction:}
    Create a matrix, \( DM \), of size \( m \times m \), where each element, \( DM_{ij} \), represents the haversine distance between nodes \( e_i \) and \( e_j \).
\end{enumerate}

\textbf{Algorithm Overview:}
\begin{enumerate}
    \item Extract coordinates of unobserved nodes.
    \item Compute node combinations.
    \item Calculate haversine distance for all node pairs.
    \item Construct the distance matrix.
\end{enumerate}

\vspace{6pt}
\hrule width \columnwidth
\section*{Cluster Disconnected Events}

\textbf{Notations:}
\begin{itemize}
    \item \( UE = \{e_1, e_2, \ldots, e_m\} \) be a set of unobserved event nodes.
    \item \( DM \) is a distance matrix, where \( DM_{ij} \) represents the distance between nodes \( e_i \) and \( e_j \).
    \item \( r \) is a radius threshold.
    \item \( n \) is the desired number of densest clusters to return.
\end{itemize}

\textbf{Methodology:}
\begin{enumerate}
    \item \textit{Identifying Nodes within Radius:}
    Determine the nodes within radius \( r \) of each other using the distance matrix \( DM \).
    \[
    N(i) = \{j : DM_{ij} \leq r\}
    \]

    \item \textit{Creating Clusters:}
    Form clusters, \( C \), based on the proximity defined above.
    \[
    C_i = \{e_j : j \in N(i)\}
    \]

    \item \textit{Sorting Clusters by Density:}
    Sort clusters based on their size (density) in ascending order.
    \[
    C_{\text{sorted}} = \text{sort}(C, \text{key} = |C_i|)
    \]

    \item \textit{Identifying Densest Clusters:}
    Identify the densest clusters, ensuring that a denser cluster does not contain the centroid of a less dense cluster.
    Let \( D_C \) be the set of densest clusters:
    \[
    D_C = \left\{C_i \in C_{\text{sorted}} : C_i \text{ is maximal dense}\right\}
    \]
    where "maximal dense" means that there is no denser cluster that contains the centroid of \( C_i \).

    \item \textit{Selecting Top \( n \) Clusters:}
    Select the top \( n \) clusters from \( D_C \).
\end{enumerate}

\textbf{Algorithm Overview:}
\begin{enumerate}
    \item Identify nodes within radius \( r \).
    \item Form clusters based on node proximity.
    \item Sort clusters by density.
    \item Identify maximal dense clusters.
    \item Select top \( n \) clusters.
\end{enumerate}

\vspace{6pt}
\hrule width \columnwidth
\section*{Add New Observers}

\textbf{Notations:}
\begin{itemize}
    \item \( UE \): Set of unobserved event nodes.
    \item \( r \): Radius threshold, consistent with prior sections.
    \item \( n \): Number of clusters to identify.
    \item \( DM \): Distance matrix among unobserved nodes.
    \item \( C \): Set representing the densest clusters.
\end{itemize}

\textbf{Methodology and Algorithm Overview:}
\begin{enumerate}
    \item \textit{Unobserved Distances:}
    Compute the pairwise distances among all unobserved event nodes within a given radius \( r \) and represent them in a distance matrix \( DM \).
    \[
    DM = \text{unipartite\_distance\_matrix}(UE, r)
    \]
    where \(\text{unipartite\_distance\_matrix}(\cdot, \cdot)\) is a function that returns a distance matrix, calculating the pairwise distances between all points in set \( UE \) within radius \( r \).

    \item \textit{Retrieve Densest Clusters:}
    Identify the \( n \) densest clusters among the unobserved event nodes \( UE \), utilizing the precomputed distance matrix \( DM \) and within the radius \( r \).
    \[
    C = \text{cluster\_disconnected}(DM, UE, r, n)
    \]
    where \(\text{cluster\_disconnected}(\cdot, \cdot, \cdot, \cdot)\) is a function that returns the \( n \) densest clusters from the unobserved events nodes \( UE \), using the precomputed distance matrix \( DM \) and within radius \( r \).
\end{enumerate}

\renewcommand*\descriptionlabel[1]{\hspace\labelsep\normalfont #1:}

\section{Datasets}
This research revolves around constructing a STROOBnet to model relationships between events from CFS that are violent, and the location of crime cameras in New Orleans. This STROOBnet comprises observer nodes (symbolizing crime cameras) and observable nodes (representing violent event locations). Edges in the STROOBnet are created by pairing each CFS event node with the closest crime camera node based on spatial range.

\subsection{Observer Node Set}

The observer node set comprises RTCC camera nodes utilized to monitor events in New Orleans. These nodes are either stationary or discretely mobile, meaning their location might instantaneously change between snapshots. Each observer node is characterized by the following properties:
\begin{itemize}
    \item \textbf{id:} Unique identifier number for each node.
    \item \textbf{membership:} Indicates if the node is a City-owned asset or a Private-owned asset.
    \item \textbf{data source:} Entity that operates the RTCC camera.
    \item \textbf{mobility:} Specifies if the node is stationary or mobile.
    \item \textbf{address:} Provides the street address of the RTCC camera.
    \item \textbf{geolocation:} Indicates the latitude and longitude of the platform.
    \item \textbf{x,y:} Represents the location in EPSG: 3452.
\end{itemize}

\subsection{Observable Node Set}
CFS events, reported to the NOPD, serve as the observable nodes in the STROOBnet, indicating violent event locations. These events undergo specific filtering based on type and location, ensuring that the observable nodes represent violent events such as homicides, non-fatal shootings, armed robberies, and aggravated assaults within the targeted region. Each CFS event node possesses several properties:

\begin{itemize}
    \item \textbf{Identification}: Each event is uniquely identified using the ``NOPD Item''.
    
    \item \textbf{Nature of Crime}: The ``Type'' and ``Type Text'' denote the nature of the CFS event.
    
    \item \textbf{Geospatial Information}: The event's location is determined by ``MapX'' and ``MapY'' coordinates, supplemented by the ``Geolocation'' (latitude and longitude), ``Block Address'', ``Zip'', and ``Police District''. 
    
    \item \textbf{Temporal Information}: Key timestamps associated with the event include ``Time Create'' (creation time in CAD), ``Time Dispatch'' (NOPD dispatch time), ``Time Arrive'' (NOPD arrival time), and ``Time Closed'' (closure time in CAD).
    
    \item \textbf{Priority}: Events are prioritized as ``Priority'' and ``Initial Priority'', ranging from 3 (highest) to 0 (none).
    
    \item \textbf{Disposition}: The outcome or status of the event is captured by ``Disposition'' and its descriptive counterpart, ``DispositionText''.
    
    \item \textbf{Location Details}: The event's specific location within the Parish is designated by ``Beat'', which indicates the district, zone, and subzone.
\end{itemize}

\section{Experimental Setup}

Reproducibility and accessibility are paramount in our experimental approach. Thus, experiments were conducted on Google Colaboratory, chosen for its combination of reproducibility—via easily shareable, browser-based Python notebooks—and computational robustness through GPU access, particularly employing a Tesla T4 GPU to leverage CUDA-compliant libraries. The software environment, anchored in Ubuntu 20.04 LTS and Python 3.10, utilized modules such as cuDF, Dask cuDF, cuML, cuGraph, cuSpatial, and CuPy to ensure computations were not only GPU-optimized but also efficient. The availability of code and results is ensured through the platform, promoting transparent and repeatable research practices.

\section{Results}
The insights from the research, visualized through graphs, heatmaps, and network diagrams, delineate the foundational attributes and effectiveness of the STROOBnet, providing a baseline for analyses and comparison with state-of-the-art approaches.

\subsection{Initial STROOBnet}
The initial state of the STROOBnet, before any methods were applied, is characterized using various visualization techniques as described below.

\begin{figure}[h!]
    \centering
    \includegraphics[width=0.95\columnwidth]{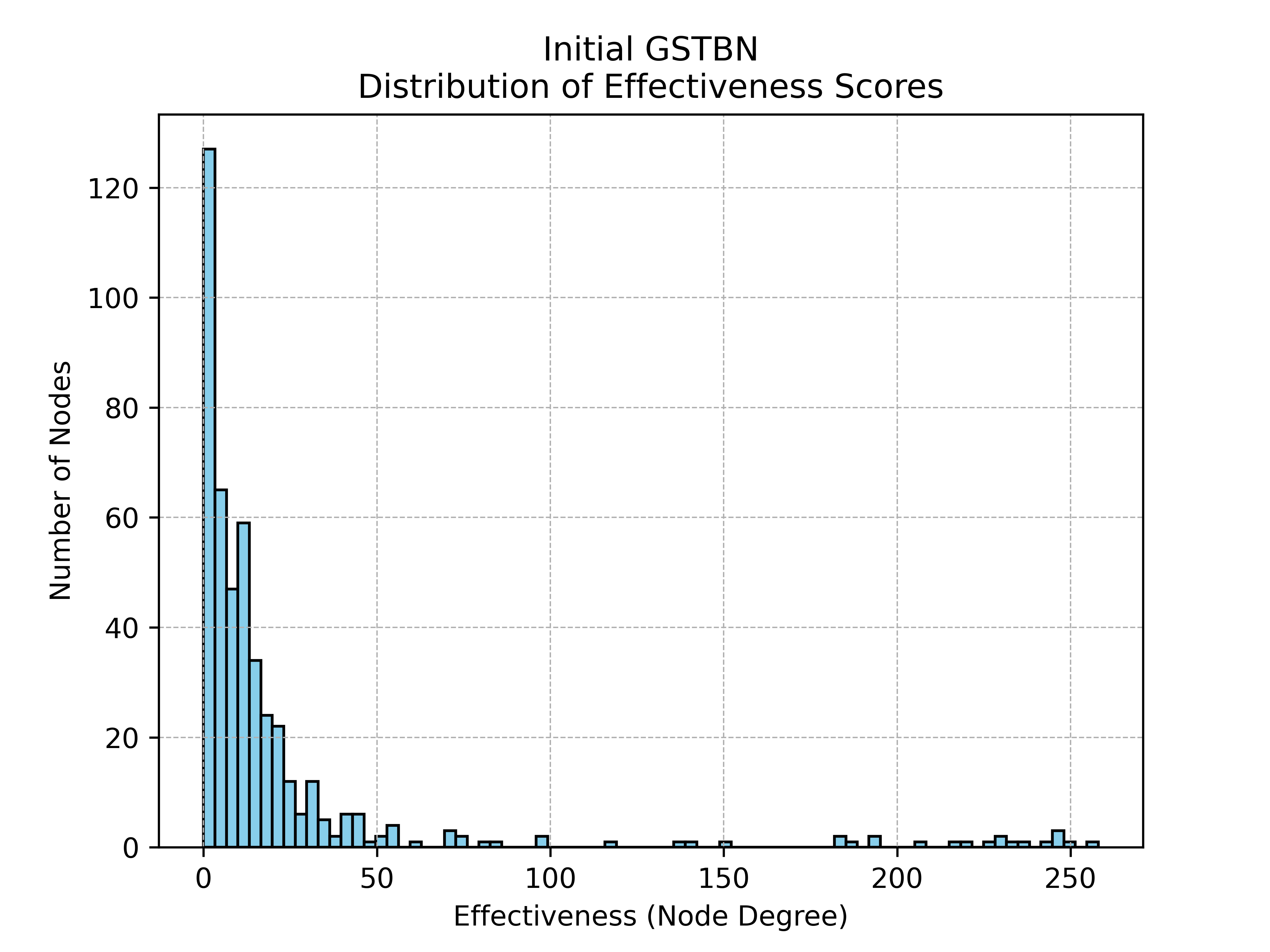}
    \caption{Histogram of effectiveness scores (node degrees) for observer nodes in the initial STROOBnet, demonstrating a power-law distribution. The x-axis represents the degree count, signifying the ability of an observer node to detect events within a specified radius, while the y-axis shows the count of nodes for each degree.}
    \label{fig:degree_distro}
\end{figure}

\begin{figure}[h!]
    \centering
    \includegraphics[width=0.95\columnwidth]{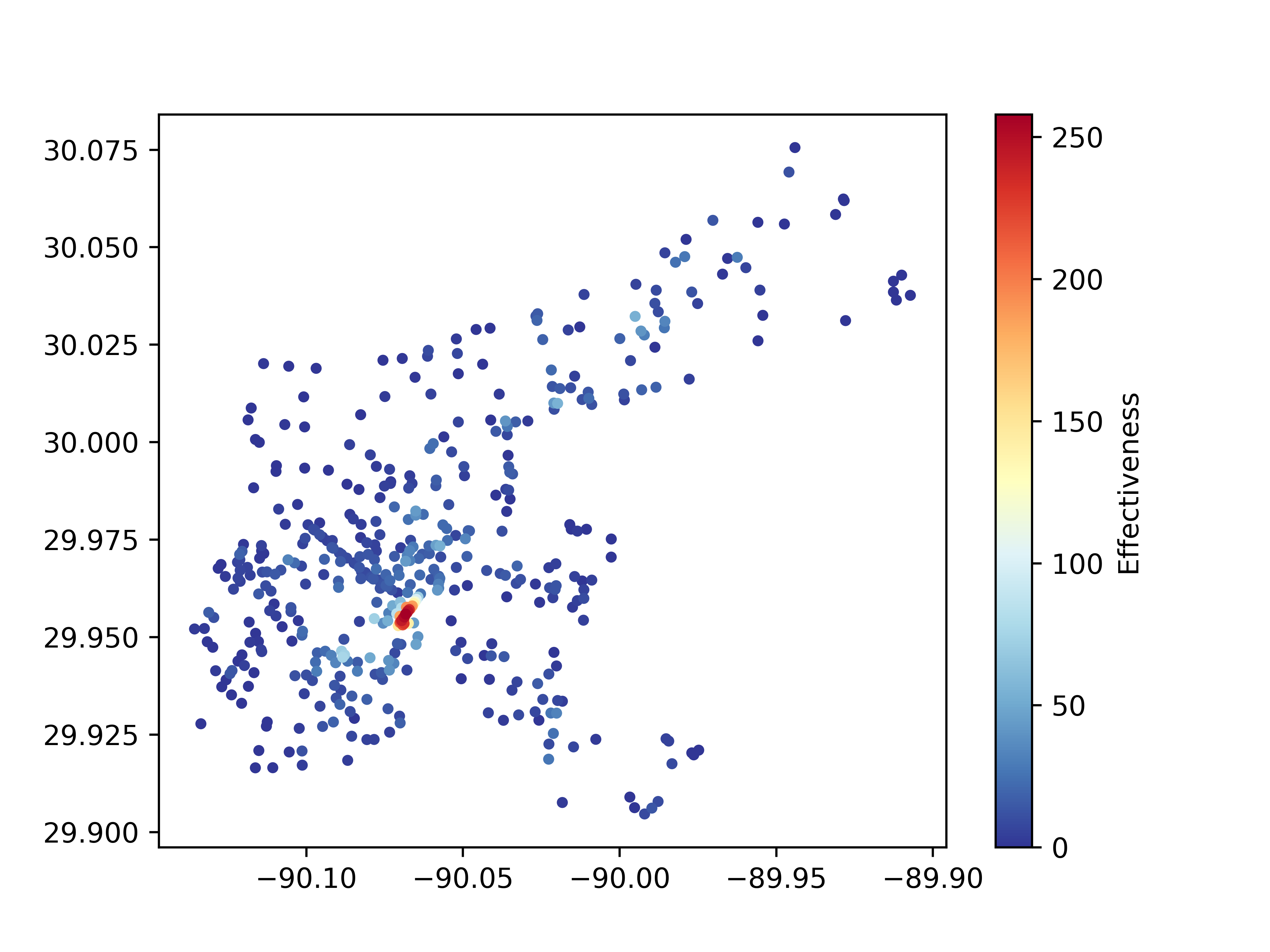}
    \caption{Heatmap of observer node effectiveness scores in the initial STROOBnet. Node spatial locations are plotted with color intensity indicating degree centrality and a gradient from blue (low effectiveness) to red (high effectiveness).}
    \label{fig:effectiveness_heatmap}
\end{figure}

\begin{figure}[h!]
    \centering
    \includegraphics[width=\columnwidth]{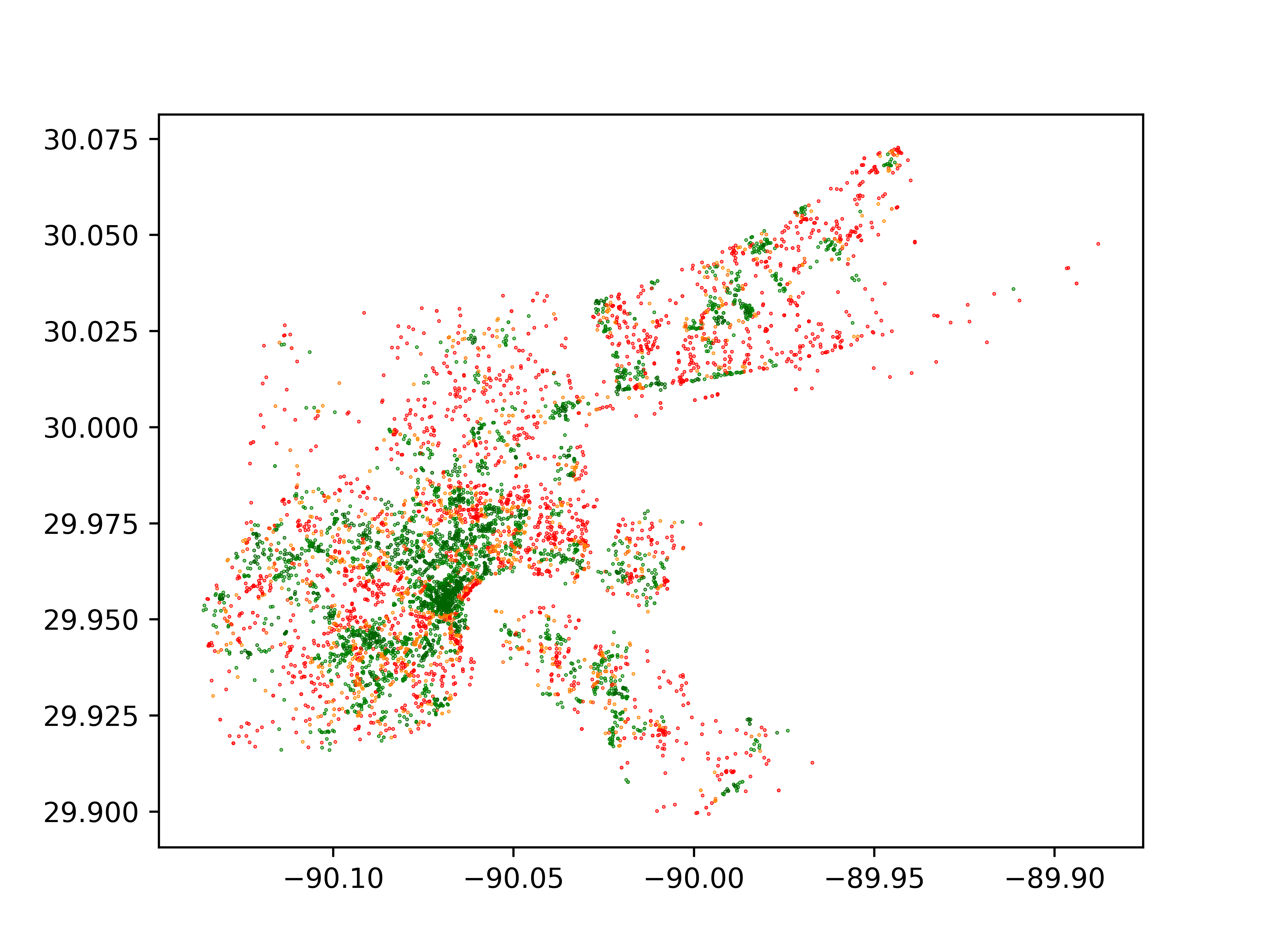}
    \caption{Spatial distribution and observational coverage of events in the initial STROOBnet. Event locations are color-coded to indicate observational coverage: deep green for multiple observers, green for a single observer, varied colors for near-observers, and red for unobserved events.}
    \label{fig:initial_gstbn_event_coverage}
\end{figure}

\subsection{Proposed Method Results: Proximal Recurrence}
The integration of 100 new nodes into the STROOBnet using the proximal recurrence strategy (with a specified radius of 0.2) produced both visual and quantitative outcomes, which are depicted and summarized in the following figures.

\begin{figure}[h!]
\centering
\includegraphics[width=\columnwidth]{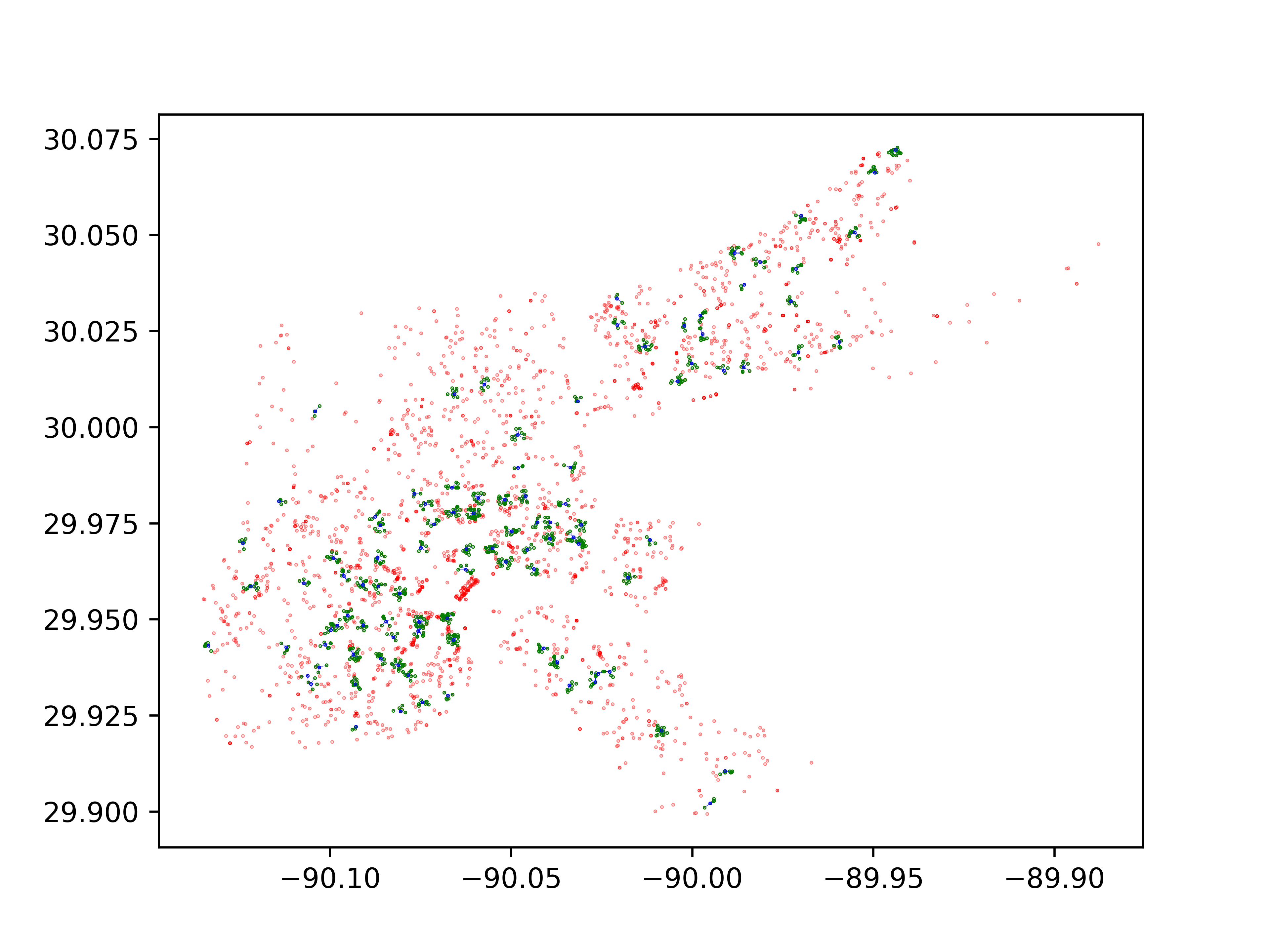}
\caption{STROOBnet visualization post proximal recurrence integration. New nodes are represented in blue, directly observed nodes in green, and unobserved nodes in red, with edges indicating observational relationships.}
\label{fig:gstbn_proximal_recurrence_subnet_with_red}
\end{figure}

\begin{figure}[h!]
\centering
\includegraphics[width=0.75\columnwidth]{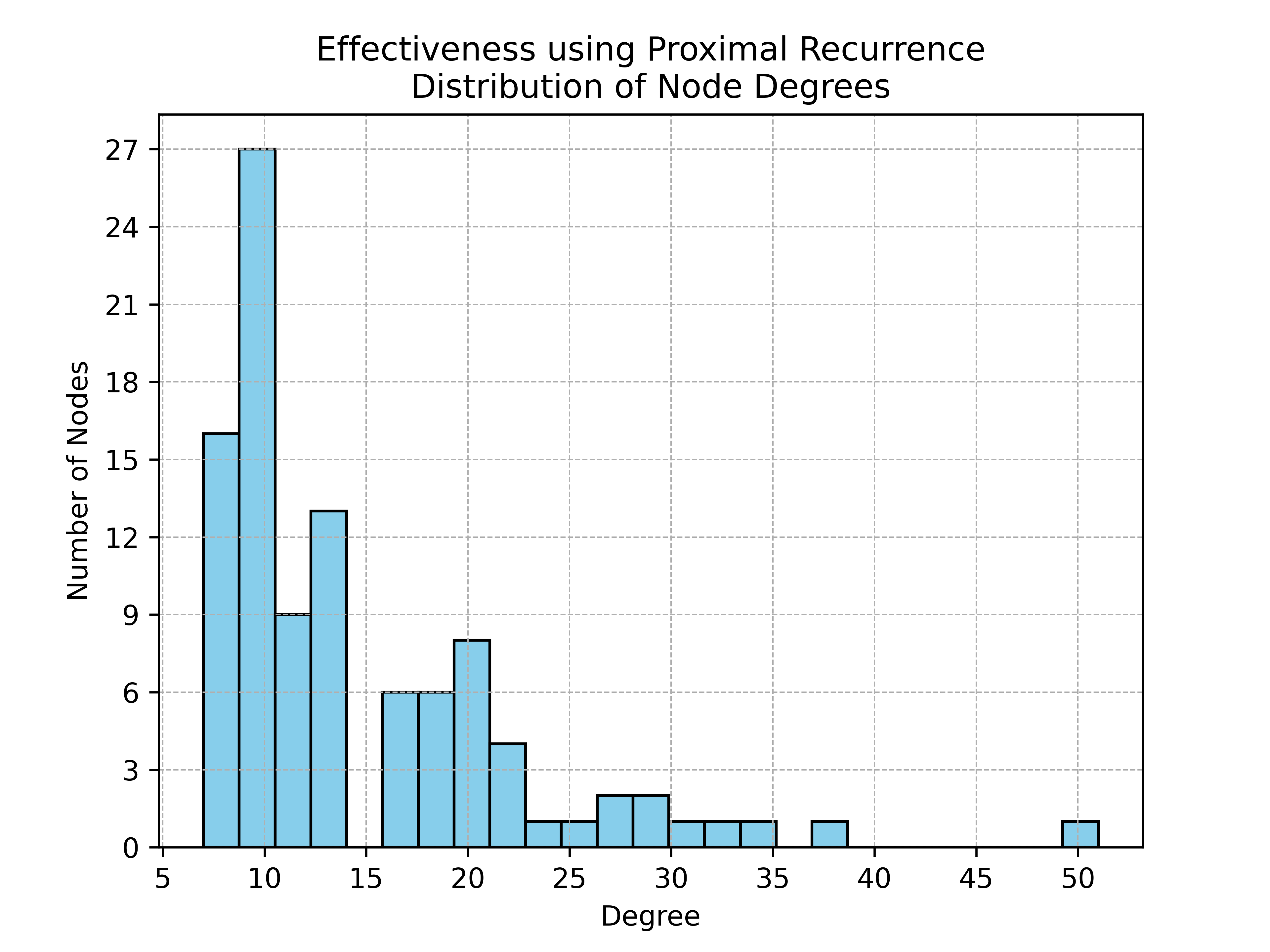}
\caption{Histogram of node degree distribution for the new nodes introduced through the proximal recurrence strategy. Axes represent degree and node count, respectively.}
\label{fig:effectiveness_proximal_recurrence}
\end{figure}

\begin{figure}[h!]
\centering
\includegraphics[width=0.75\columnwidth]{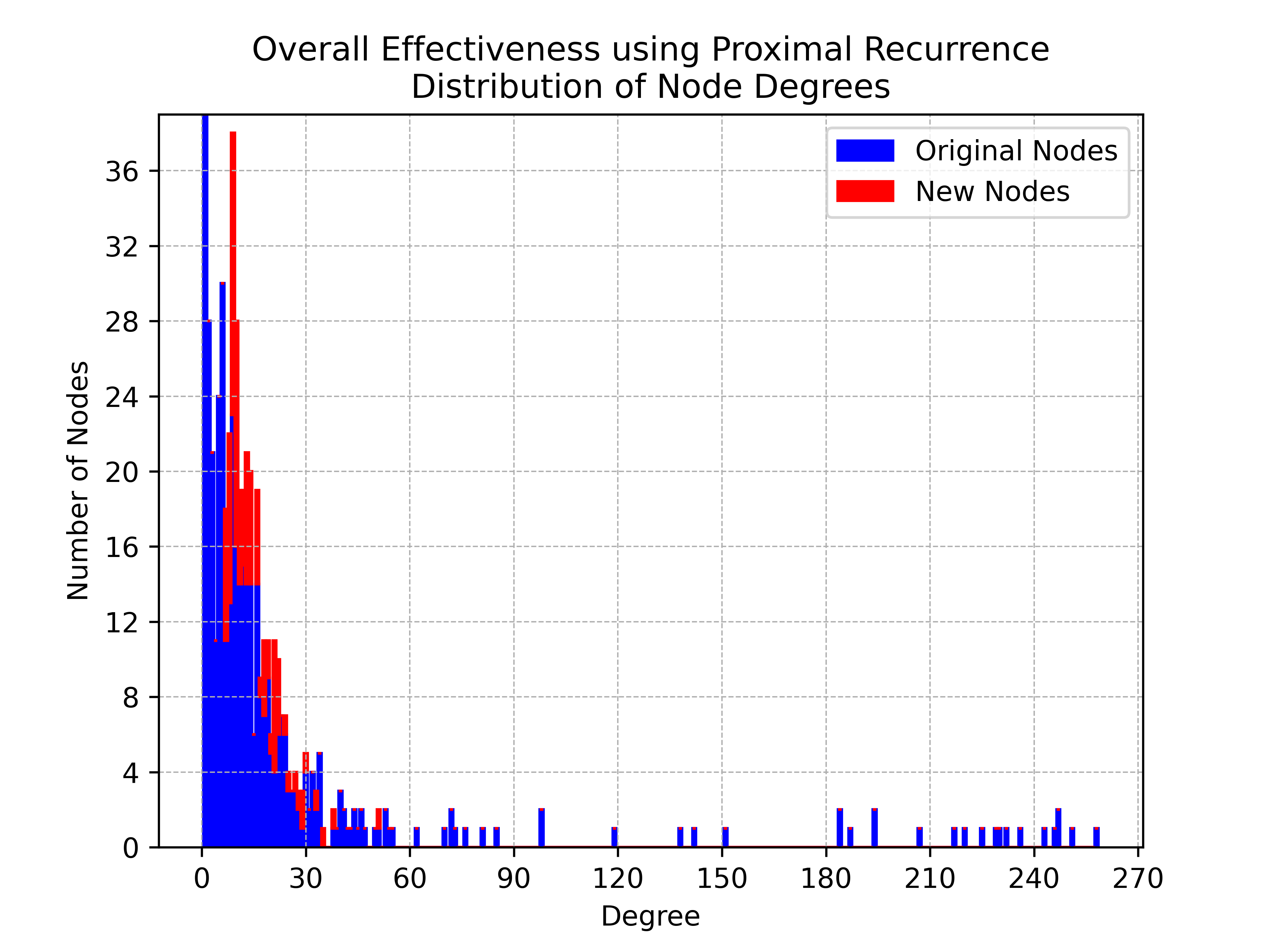}
\caption{Aggregate node degree distribution of the updated STROOBnet, with original nodes in blue and new nodes superimposed in red.}
\label{fig:overall_effectiveness_proximal_recurrence}
\end{figure}

\subsection{Comparison with Existing Methods}
This subsection presents the results of other techniques like k-means, mode, average, binning, and DBSCAN.
\\

\subsubsection{\textbf{DBScan}}
The DBScan clustering approach and its resultant effectiveness are visualized and analyzed below through various graphical representations and histograms.

\begin{figure}[h!]
    \centering
    \includegraphics[width=\columnwidth]{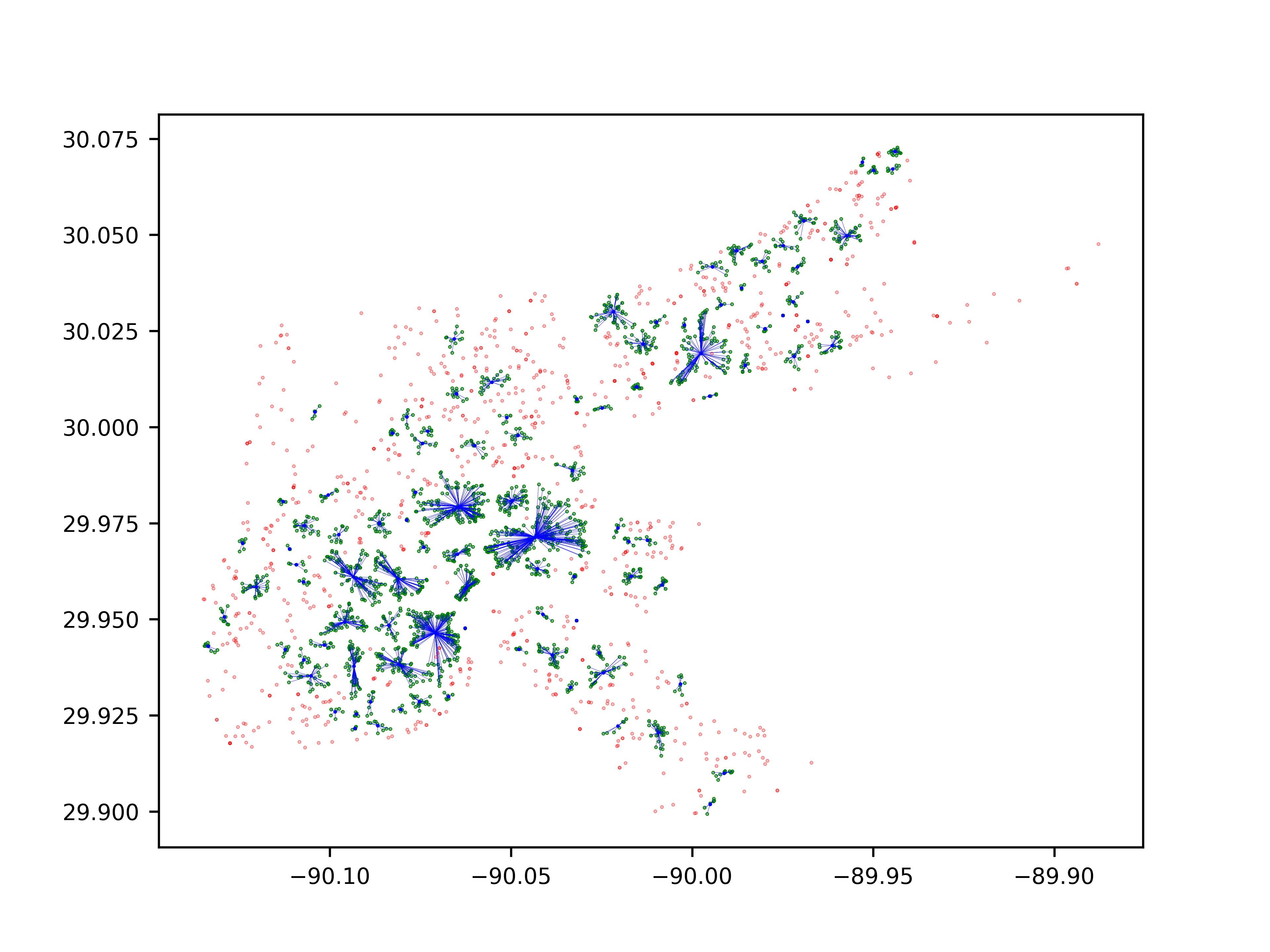}
    \caption{Centroids and associated events determined using DBSCAN clustering on unwitnessed events within the STROOBnet. Centroids (blue), clustered nodes (green), and unclustered nodes (red) are depicted, with edges connecting centroids and respective nodes.}
    \label{fig:dbscan_centroids_nonrange}
\end{figure}

\begin{figure}[h!]
    \centering
    \includegraphics[width=\columnwidth]{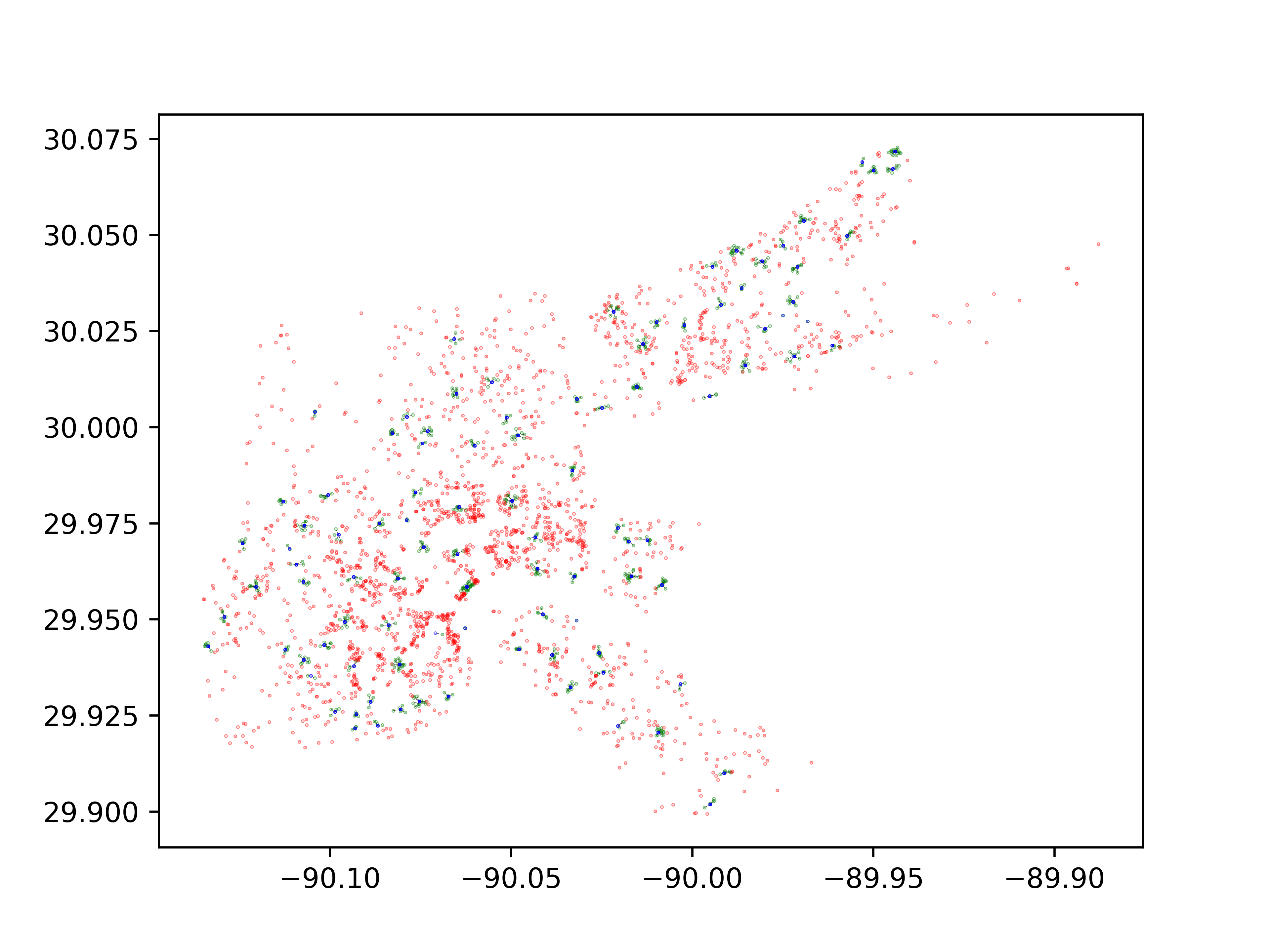}
    \caption{Node insertions based on DBSCAN clustering, emphasizing new DBSCAN nodes and filtering out events and nodes not within the centroid's spatial range.}
    \label{fig:dbscan_cluster_nonrange_gstbn}
\end{figure}

\begin{figure}[h!]
    \centering
    \includegraphics[width=0.75\columnwidth]{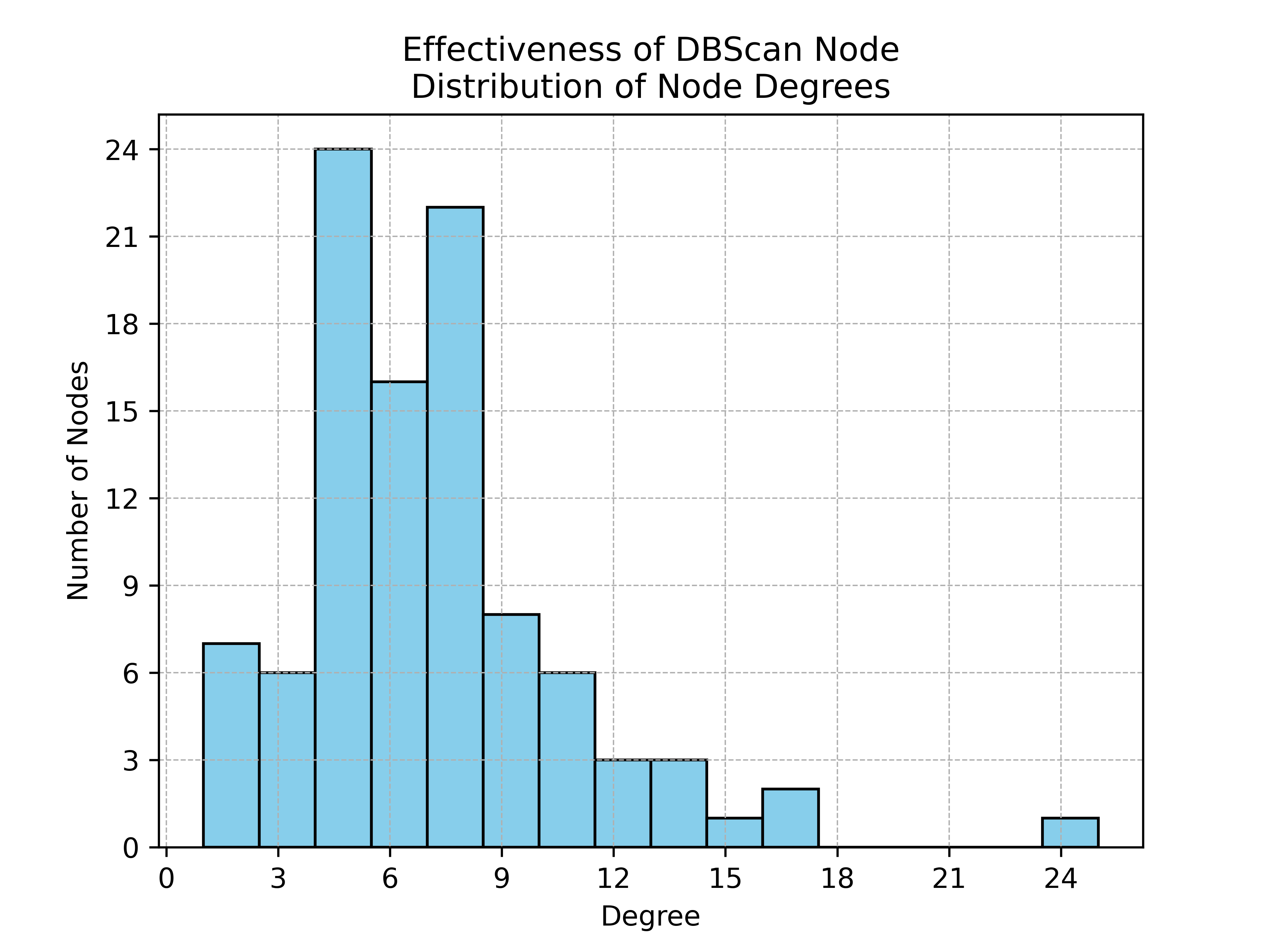}
    \caption{Distribution of node degrees for new nodes in the STROOBnet via the DBScan algorithm, with axes representing degree and node quantity.}
    \label{fig:effectiveness_dbscan_node}
\end{figure}

\begin{figure}[h!]
\centering
\includegraphics[width=0.75\columnwidth]{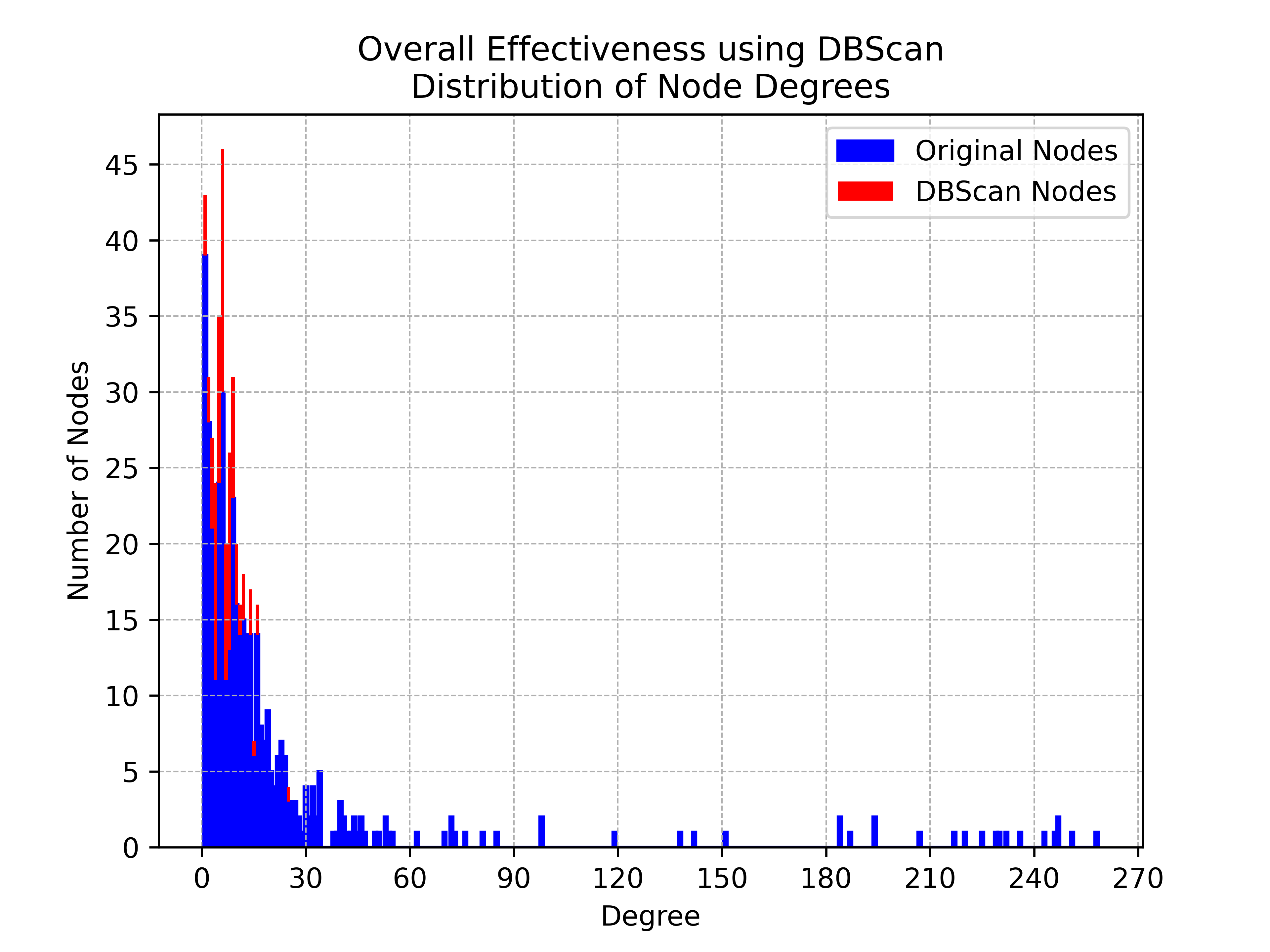}
\caption{Overall node degree distribution in the updated STROOBnet, showcasing the original distribution (blue) and the new DBScan nodes (red).}
\label{fig:overall_effectiveness_dbscan_nodes}
\end{figure}

\subsubsection{\textbf{K-means}}
The effectiveness of the K-means clustering method is illustrated through a series of visualizations and histograms.

\begin{figure}[h!]
    \centering
    \includegraphics[width=\columnwidth]{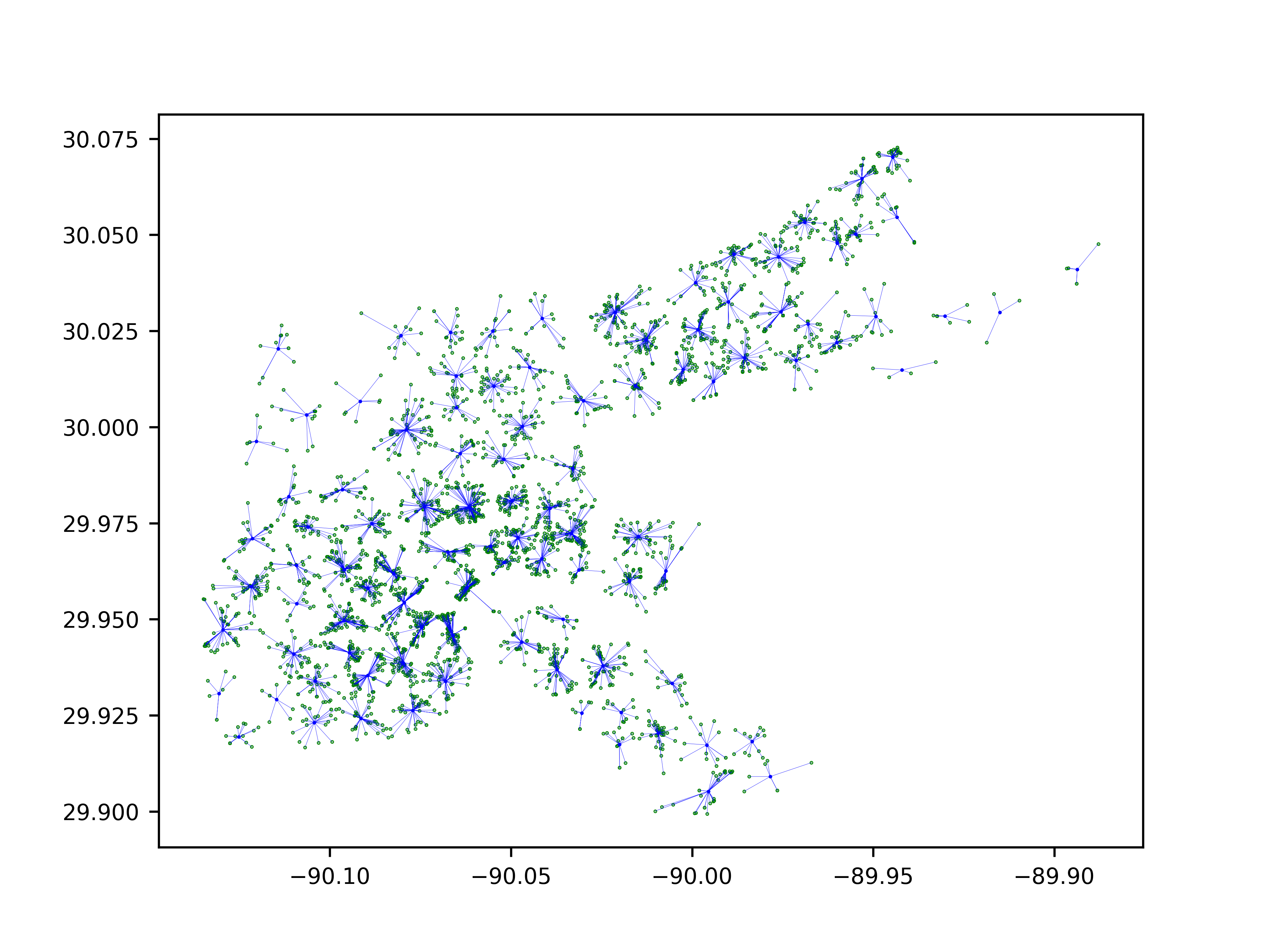}
    \caption{Visualization using K-means clustering on unwitnessed events within the STROOBnet. Centroids are in blue, clustered nodes in green, and they are interconnected with edges.}
    \label{fig:kmeans_centroids_nonrange}
\end{figure}

\begin{figure}[h!]
    \centering
    \includegraphics[width=\columnwidth]{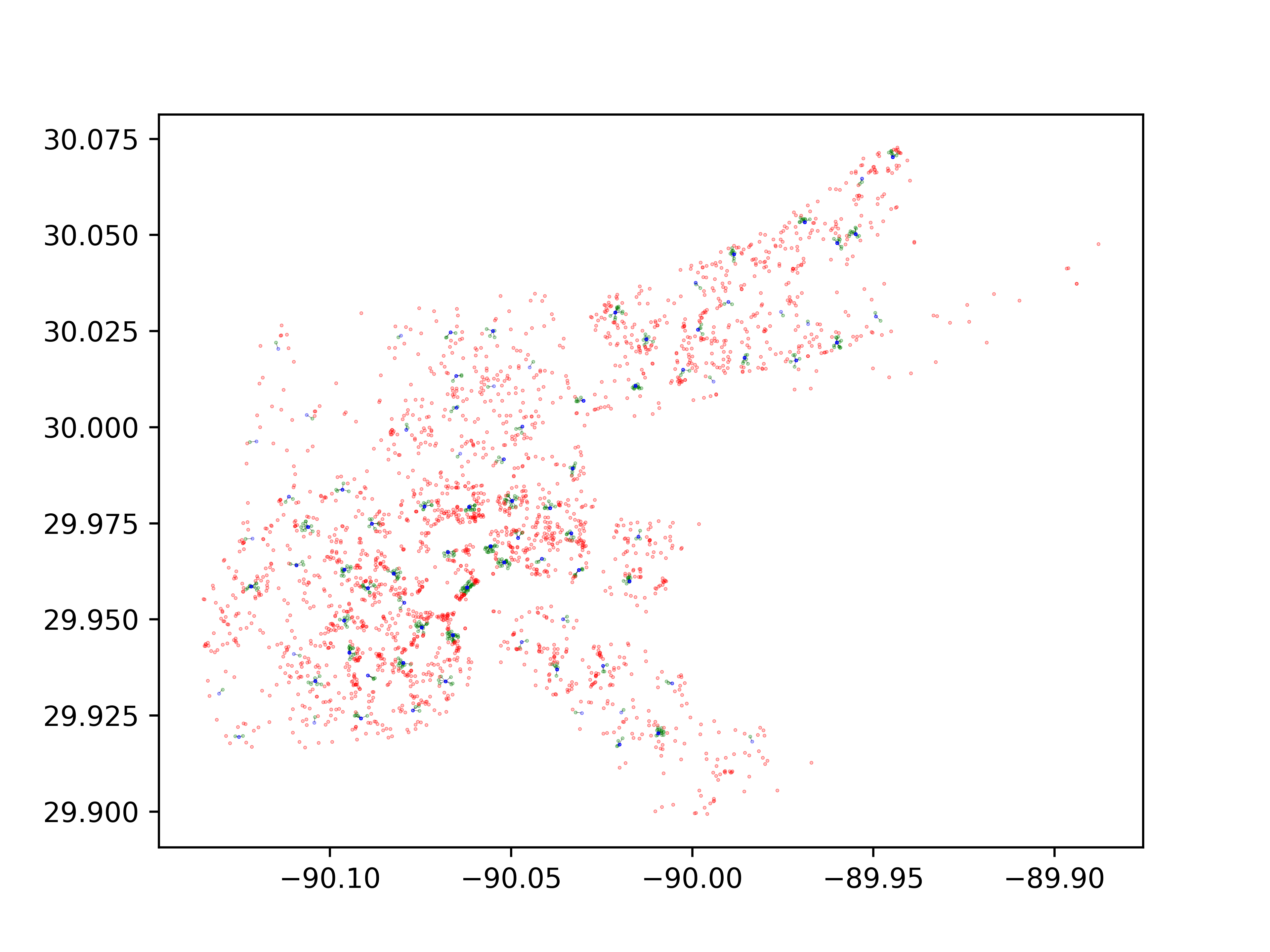}
    \caption{Centroid-based node insertions using K-means clustering, focusing on new nodes while filtering those out of the spatial range of the centroid.}
    \label{fig:kmean_cluster_nonrange_gstbn}
\end{figure}

\begin{figure}[h!]
    \centering
    \includegraphics[width=0.75\columnwidth]{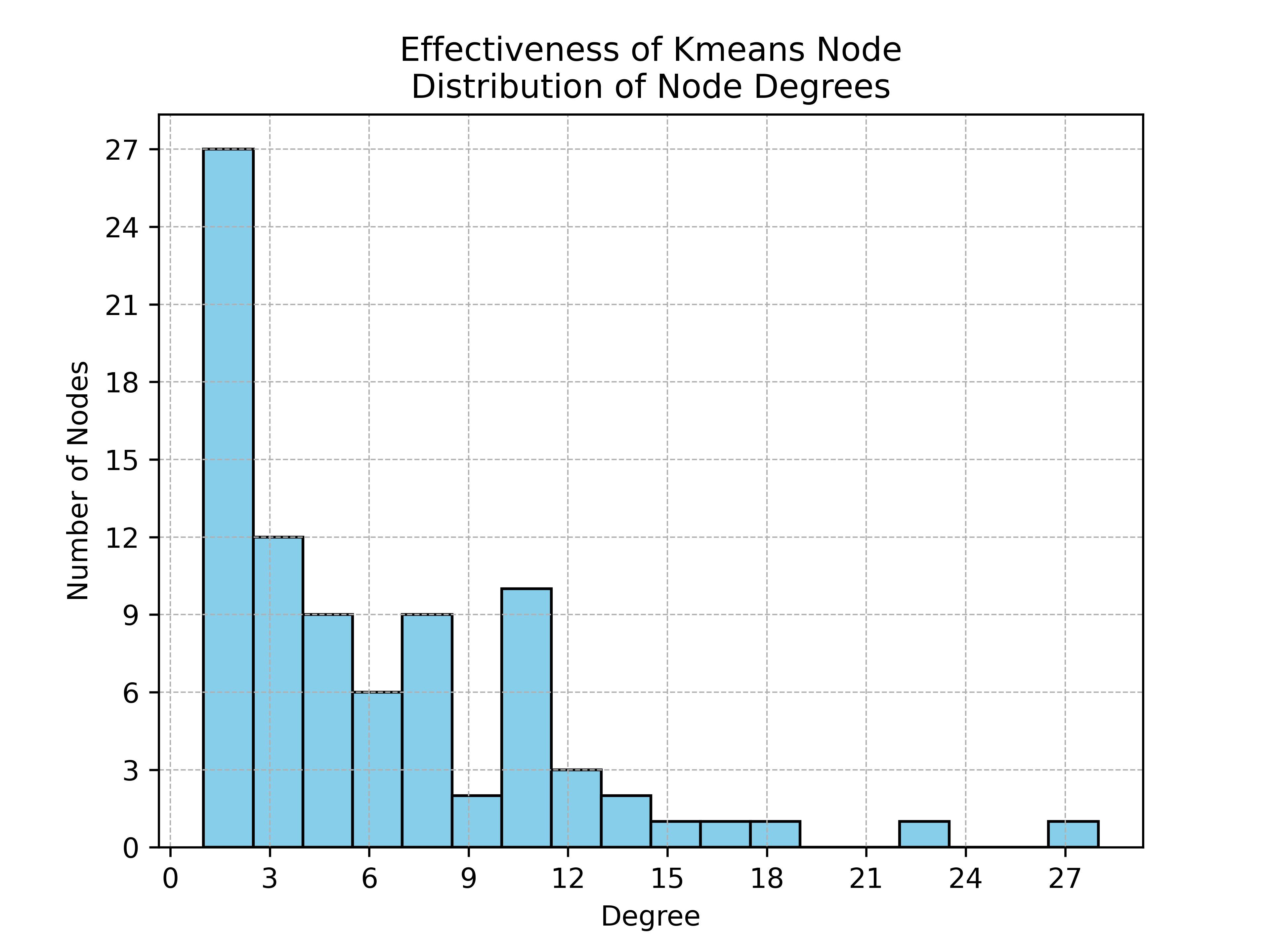}
    \caption{Histogram showing the distribution of node degrees for nodes added through the K-means method.}
    \label{fig:effectiveness_kmeans_node}
\end{figure}

\begin{figure}[h!]
\centering
\includegraphics[width=0.75\columnwidth]{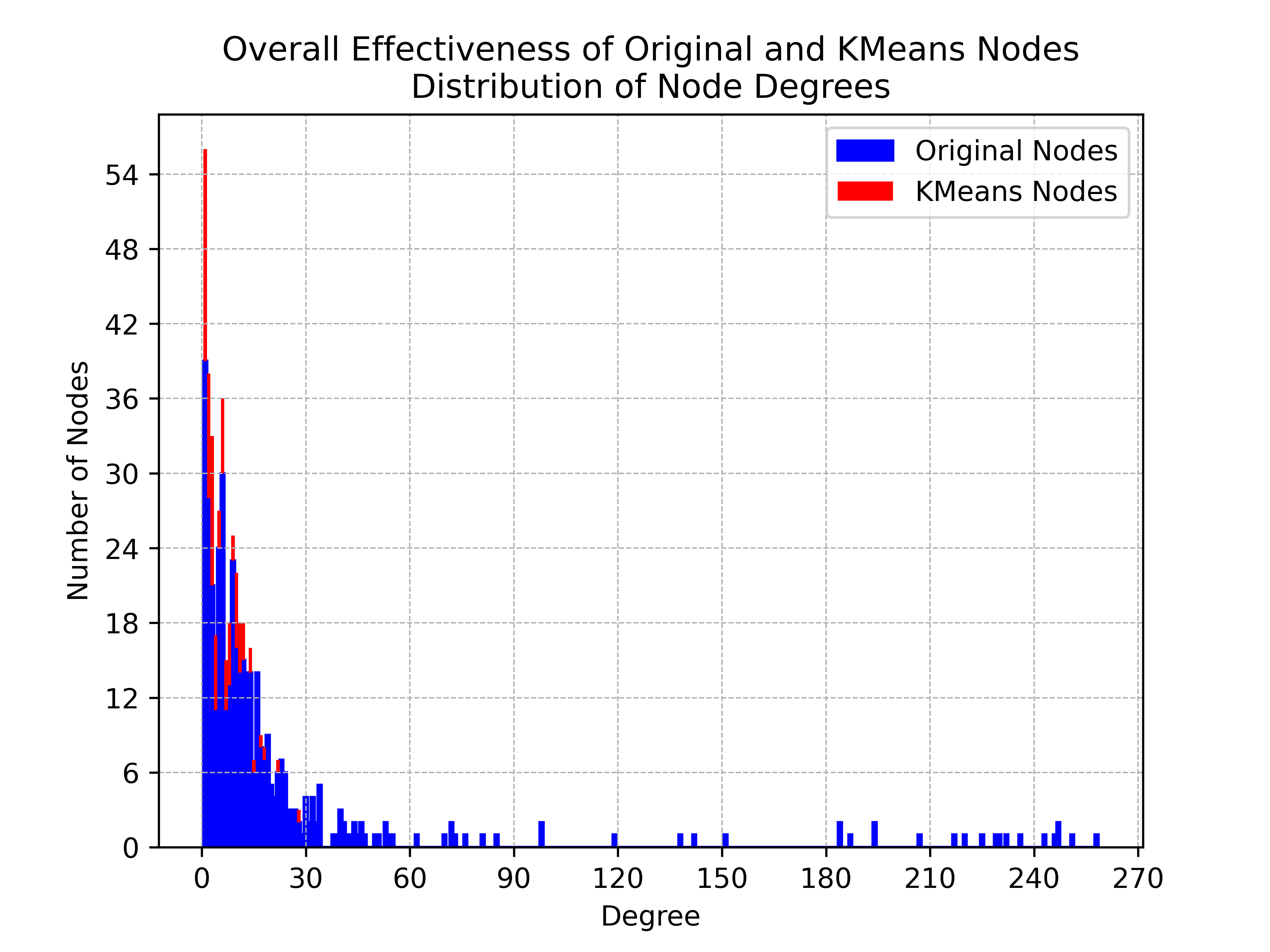}
\caption{Combined histogram indicating the node degree distribution in the updated STROOBnet, distinguishing between original (blue) and K-means added nodes (red).}
\label{fig:overall_effectiveness_kmeans_nodes}
\end{figure}

\subsubsection{\textbf{Mode Clustering}}
The mode clustering technique is demonstrated using various graphical representations and data distribution plots.

\begin{figure}[h!]
\centering
\includegraphics[width=\columnwidth]{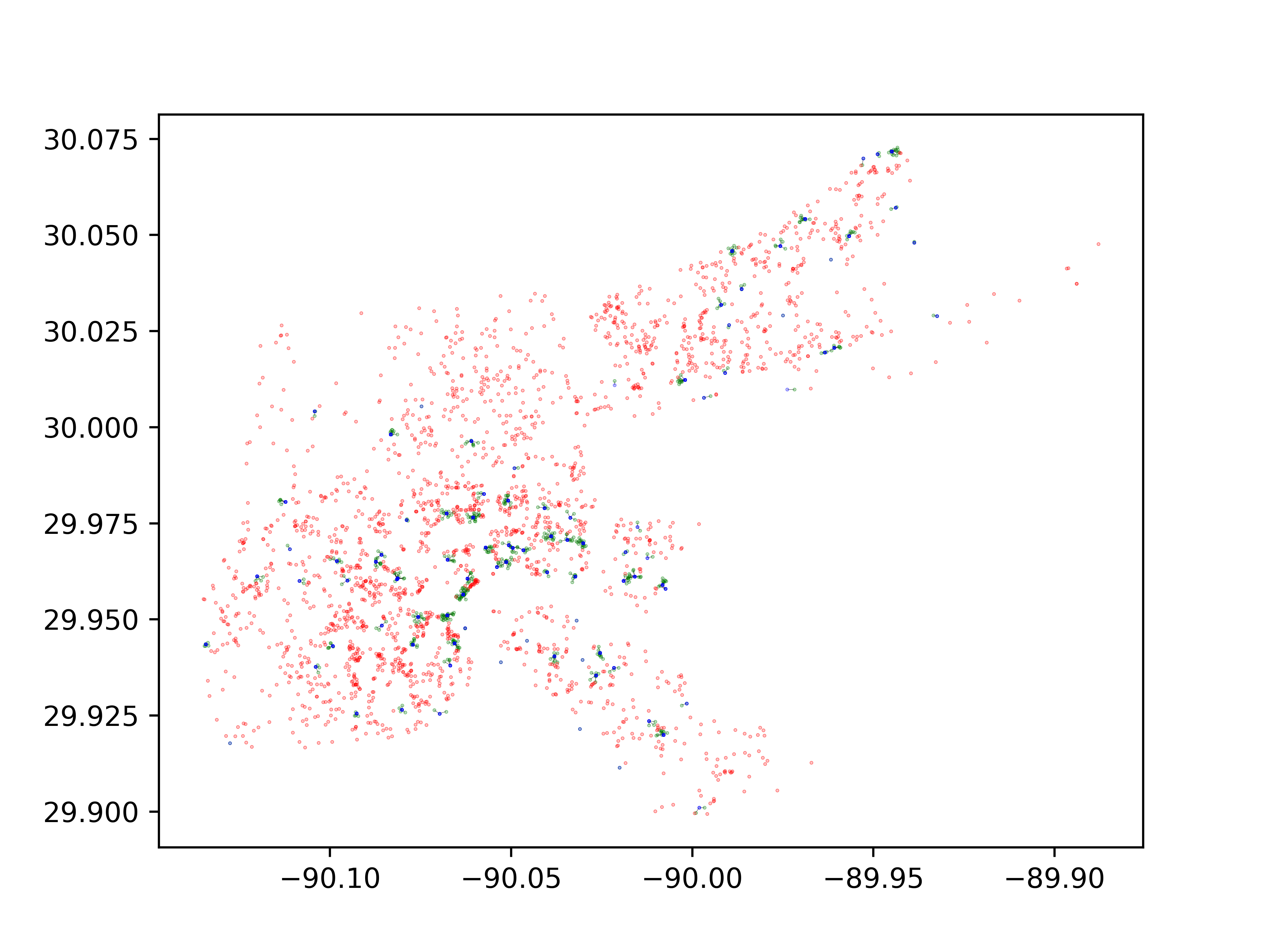}
\caption{STROOBnet utilizing the Mode-adjusted variant of K-means approach, featuring new nodes (blue), directly observed nodes (green), and remaining unobserved nodes (red).}
\label{fig:mode_standard_gstbn_unobserved}
\end{figure}

\begin{figure}[h!]
\centering
\includegraphics[width=0.75\columnwidth]{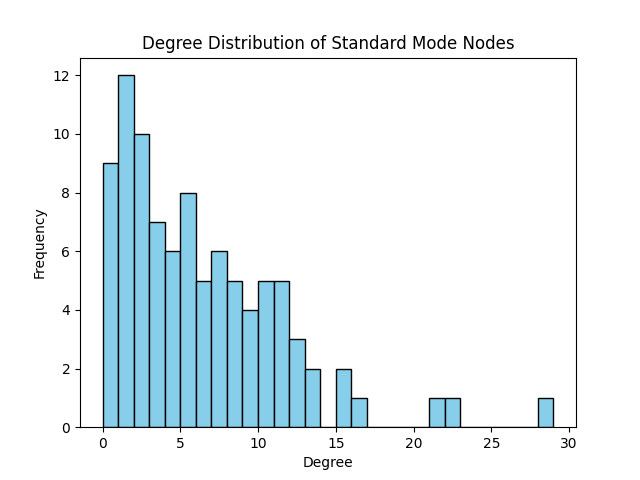}
\caption{Histogram indicating the distribution of node degrees for nodes added via the Standard Mode strategy.}
\label{fig:effectiveness_mode_standard_nodes}
\end{figure}

\begin{figure}[h!]
\centering
\includegraphics[width=0.75\columnwidth]{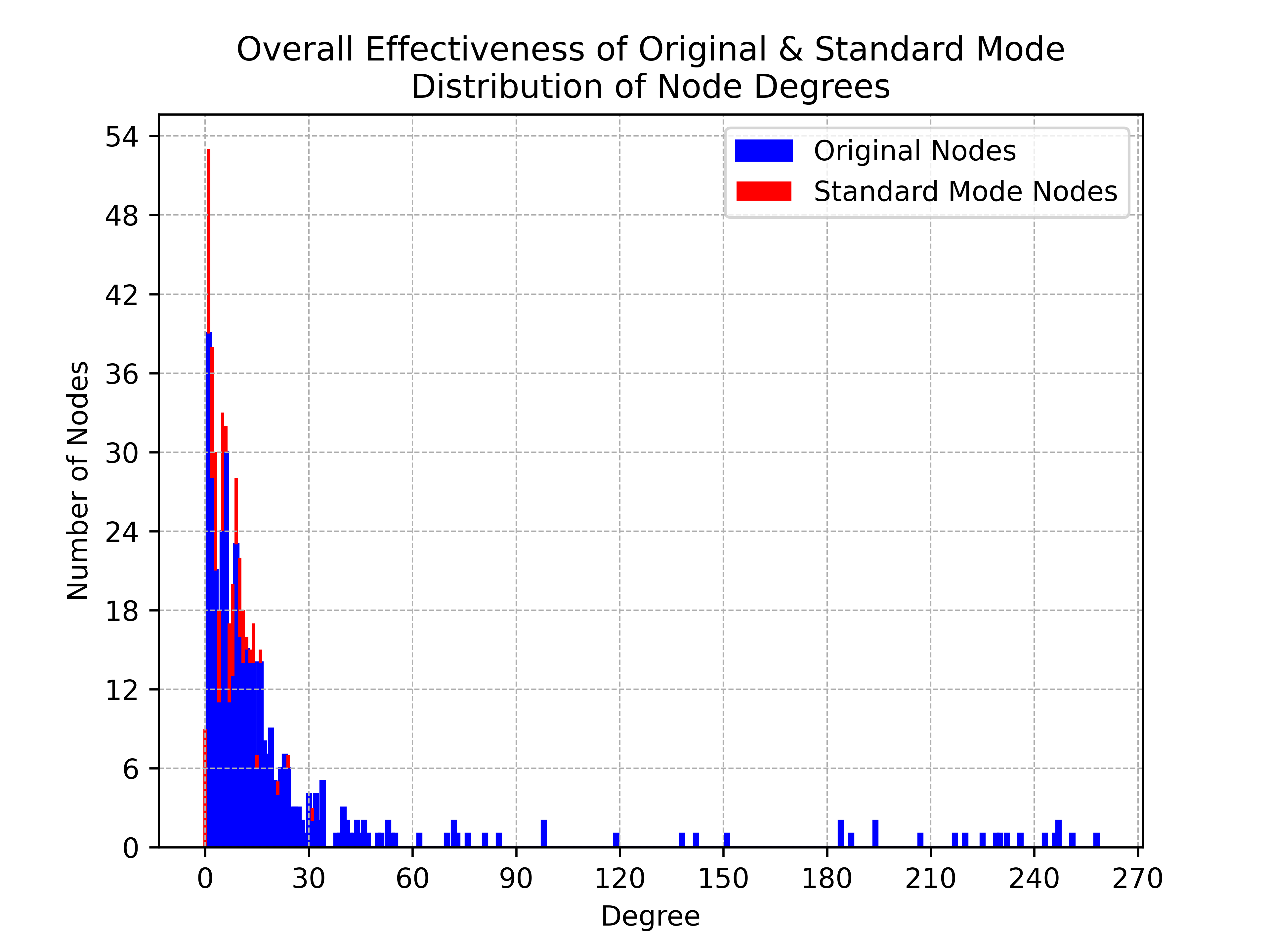}
\caption{Histogram of node degree distribution in the STROOBnet, illustrating both original (blue) and Standard Mode added nodes (red).}
\label{fig:overall_effectiveness_mode_kmeans_nodes}
\end{figure}

\section{Discussion}

\subsection{Evaluating Effectiveness}

\subsubsection{Degree Centrality}
Utilizing a bipartite behavior, the network connects observers to observable events through edges. The centrality of a node quantifies its efficacy in witnessing events. The distribution of node degrees provides insights into the network's capability to capture events through its observers.

\subsubsection{Desired Transformations in Degree Centrality Distributions}
The network initially exhibits a power-law distribution in degree centrality, where few nodes are highly effective and most are not. An objective is to shift this to a skewed Gaussian distribution by introducing new nodes, thereby redistributing node effectiveness and enhancing network performance. The combined histogram of original and new node centrality acts as a metric for evaluating this shift. A rightward histogram shift, visible in the results, indicates enhanced network performance due to the inclusion of new nodes with higher degree centrality. This transformation and the resulting outcomes set a benchmark for evaluating different clustering strategies in subsequent sections.

\subsection{Comparative Analysis of Clustering Approaches}

\subsubsection{Limitations of Conventional Clustering Strategies}
Conventional strategies like k-means and DBSCAN manifest limitations in controlling cluster diameter and centroid assignment, which can obscure genuine data patterns and hinder the transition from a power-law to a skewed Gaussian distribution upon the addition of new nodes. The proximal recurrence approach offers improved control and contextually relevant centroid assignment, addressing these issues.

\subsubsection{The Challenge of Centroid Averaging}
In domains such as crime analysis, centroid averaging, or calculating a point that represents the mean position of all cluster points, can misrepresent actual incident locations in spatially significant contexts. Traditional centroid assignment can produce misplaced centroids, leading to inaccurate analyses and strategies, such as incorrectly identifying crime hot spots. The proximal recurrence method ensures accurate data representation by managing centroid assignments and controlling cluster diameters, particularly in contexts requiring precise spatial data point accuracy.

\subsection{Inadequacies of Mode-Based Analysis}

\subsubsection{Neglect of Spatial Clustering}
Mode-based analysis effectively identifies high-recurrence points but neglects the spatial clustering of nearby values. This oversight can miss opportunities to place observer nodes in locations where a cluster of neighboring points within a specific range might yield higher effectiveness than a singular high-incidence point.

\subsubsection{Lack of Comprehensive Insights}
While mode-based analysis hones in on recurrence, it disregards insights from considering spatial and neighboring data, identifying locations with high incidents but failing to define a cluster shape. Consequently, it misses locations near points of interest, potentially omitting optimal node identification. The approach discussed counters some of these deficiencies by incorporating both recurrence and proximate incidents into its analysis, offering a more balanced view.

\subsection{Advantages of Proximal Recurrence Approach}

\subsubsection{Harmonizing Recurrence and Spatiality}
The proximal recurrence approach integrates incident recurrence and spatial analysis, ensuring a comprehensive data perspective. It does not only identify high-incident locations but also accounts for spatial contexts and neighboring events, safeguarding against isolated data interpretation and enhancing the analysis's comprehensiveness.

\subsubsection{A Middle Ground: Specificity vs. Generalization}
The approach balances specificity and generalization by identifying incident points and considering their spatial contexts. It prevents the dilution of specific data points in generalized clustering and avoids omitting relevant neighboring data in specific point analysis. It thereby ensures the analysis remains accurate and prevents omission of pivotal data points.

\subsubsection{Defined Constraints for Enhanced Performance}
The integration of defined constraints, such as a predetermined cluster diameter, optimizes performance and ensures contextually relevant insights. By adhering to predefined limits, the approach maintains a balanced analysis, ensuring insights are precise and applicable to real-world scenarios.

\subsection{Comparison with Histogram Approach}

\subsubsection{Hyperparameter Tuning and Large Datasets}
Histograms require tuning of the bin number, a task that can be particularly intricate for large datasets due to its impact on analysis and result quality. Conversely, the proximal recurrence approach demands no hyperparameter tuning, providing straightforward applicability without iterative refinement.

\subsubsection{Accuracy, Precision, and Flexibility}
Histograms may lose fine data structures by aggregating data into bins. In contrast, the proximal recurrence approach evaluates points on actual distances, capturing accurate clustering patterns and enabling identification of non-rectangular clusters, thereby providing a more realistic data pattern representation.

\subsubsection{Adaptability and Relationship Consideration}
While histograms use uniform bin sizes, potentially misrepresenting datasets with varied density regions, the proximal recurrence approach adapts to different densities by evaluating proximity, not a fixed grid, and utilizes explicit pairwise distances, ensuring direct measurement of point relationships.

\subsubsection{Customizability and Consistency}
While histograms offer bin size as the primary customizable parameter, the proximal recurrence approach provides parameters, such as distance thresholds, for tuning based on data and problem specifics, and ensures consistent results, unaffected by alignment considerations.

\section{Conclusion}

This study delved into the intricacies of STROOBnet, focusing on node insertions and clustering methodologies. The aim was to understand the implications of different strategies on the network's performance, particularly regarding the efficacy of observer nodes in event detection. From an initial power-law distribution in node degree centrality, the study sought strategies that could enhance uniformity and event detection across STROOBnet.

\subsection{Key Takeaways}

\begin{enumerate}
    \item \textbf{Degree Centrality Transformation:}
    By integrating new nodes through various strategies, node degree centrality evolved towards a skewed Gaussian distribution, signifying an enhanced network-wide event detection capability.
    
    \item \textbf{Effectiveness of Clustering Techniques:}
    Traditional methods like k-means had limitations, while the proximal recurrence approach emerged superior in centroid assignment and optimizing cluster dimensions.
    
    \item \textbf{Mode-Based Insights:}
    The standard mode approach was proficient in pinpointing high-recurrence points but lacked in spatial clustering comprehension, often missing optimal node placements.
    
    \item \textbf{Excellence of Proximal Recurrence:}
    This approach effectively identified high-incident areas and emphasized the spatial context, ensuring comprehensive and relevant data analysis.
    
\end{enumerate}

\subsection{Implications}

The results emphasize the importance of a holistic clustering strategy. This strategy should discern optimal node placements, acknowledge spatial dynamics, and represent data with precision, especially in fields demanding high spatial accuracy.

\end{document}